# Topology-Aware Graph Reinforcement Learning for Dynamic Routing in Cloud Networks

Yuxi Wang
Carnegie Mellon University
Pittsburgh ,USA

Heyao Liu
Northeastern University
Boston, USA

Guanzi Yao
Northwestern University
Evanston, USA

Nyutian Long
New York University
New York, USA

Yue Kang*
Carnegie Mellon University
Pittsburgh ,USA

*Abstract*-This paper proposes a topology-aware graph reinforcement learning approach to address the routing policy optimization problem in cloud server environments. The method builds a unified framework for state representation and structural evolution by integrating a Structure-Aware State Encoding (SASE) module and a Policy-Adaptive Graph Update (PAGU) mechanism. It aims to tackle the challenges of decision instability and insufficient structural awareness under dynamic topologies. The SASE module models node states through multi-layer graph convolution and structural positional embeddings, capturing high-order dependencies in the communication topology and enhancing the expressiveness of state representations. The PAGU module adjusts the graph structure based on policy behavior shifts and reward feedback, enabling adaptive structural updates in dynamic environments. Experiments are conducted on the real-world GEANT topology dataset, where the model is systematically evaluated against several representative baselines in terms of throughput, latency control, and link balance. Additional experiments, including hyperparameter sensitivity, graph sparsity perturbation, and node feature dimensionality variation, further explore the impact of structure modeling and graph updates on model stability and decision quality. Results show that the proposed method outperforms existing graph reinforcement learning models across multiple performance metrics, achieving efficient and robust routing in dynamic and complex cloud networks.

## I. INTRODUCTION

With the rapid development of cloud computing and the widespread deployment of large-scale distributed systems, the demand for data communication among cloud servers has grown explosively [1]. To ensure stable operation and efficient interaction of multi-tenant services within data centers, intelligent optimization of routing strategies has become a critical issue. For example, in wide-area research networks such as GEANT, sudden increases in traffic load can cause network links to approach full capacity within a short period of time. If the system fails to reroute traffic promptly, localized congestion may rapidly propagate across interconnected links, resulting in large-scale performance degradation. This scenario underscores the importance of adaptive and intelligent routing mechanisms to mitigate congestion and sustain service reliability under dynamic workloads. Traditional static routing mechanisms cannot perceive environmental changes. As a result, they often fail to perform well in highly dynamic and heterogeneous cloud environments[2]. At the same time, the network structure among cloud servers is highly complex and diverse. It includes various hierarchical and large-scale topologies. These dynamic characteristics further increase the difficulty of routing decisions. Achieving low latency, high throughput, and load balancing in dynamic topologies remains a core challenge for current cloud network scheduling systems[3].

In recent years, graph structures have become an important tool for modeling cloud server topologies. In cloud environments, servers, switches, and links naturally form a graph. Nodes represent devices, and edges represent network connections or communication paths. This graph structure is not static. It changes over time due to task loads, link status, and fault-tolerant mechanisms [4-7]. Therefore, routing mechanisms based only on representation learning or static graphs cannot effectively capture the evolving information of the network structure. They also fail to adapt to fluctuations in real-time communication quality. Especially under large-scale deployment and multi-tenant shared resources, it is critical to extract valuable features from graph hierarchies, connection densities, and path connectivities. These features are not only fundamental for guiding intelligent routing strategies and driving the intelligent evolution of cloud networks but also play a pivotal role in advancing leading-edge applications across diverse domains. In large language models, for example, efficient management of complex graph structures and adaptive routing directly influence distributed training efficiency and model serving reliability [8-12]. In the financial sector, such feature extraction underpins secure, high-frequency transaction processing and robust risk control in dynamic market environments [13-16]. Likewise, in medical systems, adaptive feature-driven routing strategies are essential for ensuring real-time data availability, supporting critical decision-making, and enabling resilient healthcare networks [17-21]. These features can guide the generation of intelligent routing strategies and drive the intelligent evolution of cloud networks.

Reinforcement learning has emerged as an important tool in network scheduling and resource allocation due to its adaptability and capability for policy optimization in complex decision-making problems. When combined with graph neural networks, graph reinforcement learning offers a new paradigm for solving dynamic routing problems. These models can perceive both the local state and global attributes of each node in the current topology [22-23]. They can iteratively optimize routing strategies based on environmental feedback during sequential decision processes. This mechanism aligns well with the evolving nature of "state-action-reward" in cloud environments. It supports challenges such as high-dimensional state spaces, complex dependencies, and delayed feedback. By integrating topology-aware capabilities, graph reinforcement learning can precisely identify network bottlenecks and potential conflict paths. This enables the construction of more robust decision-making systems[24-26].

Although some studies have attempted to apply reinforcement learning or graph learning methods to routing optimization, most remain at the level of static graph modeling or shallow policy learning. They cannot deeply model the dynamic evolution of cloud server topologies. Furthermore, existing methods often overlook the importance of "structure awareness" for policy generalization and network stability[27]. As a result, they perform poorly in scenarios with frequent topology changes or dense task loads [28]. They also suffer from limited policy transferability and a tendency to fall into local optima [29]. These limitations hinder overall system performance. There is an urgent need for a model framework with structural adaptability, state awareness, and policy interpretability. Such a framework should maintain decision effectiveness under dynamic topologies and support intelligent scheduling in large-scale, multi-tenant, and heterogeneous environments[30]. By introducing structure-aware mechanisms, the model actively identifies critical communication paths and congested nodes in the graph. This enhances robustness and global awareness in path selection. Combined with policy gradient-based reinforcement learning algorithms, the model supports continuous learning and policy stability under rapidly changing topologies[31]. On this basis, the study promotes the evolution of graph learning paradigms toward task-driven and autonomous decision-making. It provides an intelligent and automated solution for cloud data center network scheduling. This work not only improves routing efficiency and service quality but also lays a theoretical foundation for building cloud-intelligent networks with real-time perception and adaptive optimization capabilities[32].

## II. Related work

The intersection of graph neural networks (GNNs) and reinforcement learning has driven substantial progress in adaptive policy optimization for complex, dynamic environments. Early foundational work demonstrated the ability of graph-based deep learning to learn and generate distributed routing protocols, showing that GNNs can effectively capture both local and global structural information—a methodological inspiration for structure-aware state representation in our own approach [33]. Building on this, multi-agent reinforcement learning frameworks have been developed to support adaptive resource allocation, highlighting the flexibility of policy-driven strategies in distributed systems [34]. These advances set the stage for combining structural awareness with adaptive policy updates.

Subsequent studies have refined these foundations by leveraging GNN-based collaborative perception to enhance adaptive scheduling in distributed architectures, emphasizing the critical role of high-order structural dependency modeling for efficient decision-making [35]. At the same time, techniques for entity-aware graph modeling have been introduced to improve structured information extraction, demonstrating how expressive graph-based encodings can capture nuanced relationships within complex systems—an idea that resonates with our use of multi-layer graph convolution and positional embeddings [36].

In addition, research on robust anomaly detection has explored the use of selective noise injection and feature scoring to strengthen model stability in unsupervised scenarios, offering methodological tools for enhancing decision robustness under uncertainty [37]. Complementary to this, privacy-enhanced federated learning frameworks have been proposed to address distributed optimization challenges and maintain structural adaptability in heterogeneous, evolving networks [38]. These concepts inform our design of adaptive policy updates that maintain system stability across diverse conditions.

The application of AI-driven, multi-agent scheduling has further demonstrated how scalable control can be achieved in microservice-based and cloud environments by incorporating flexible reinforcement learning policies [39]. More recently, the integration of GNNs with transformer architectures has opened new possibilities for unsupervised system anomaly discovery, blending structural modeling with temporal sequence reasoning—a synergy that supports the topology-aware mechanisms in our own model [40]. Likewise, temporal graph attention methods have been proven effective for modeling spatiotemporal dependencies in dynamic graphs, further strengthening the methodological underpinnings of our structure-aware state encoding module [41].

Expanding on these ideas, spatiotemporal feature learning networks have shown how relevant features can be adaptively extracted from sequential, high-dimensional data [42]—a principle directly relevant to both our graph representation and policy optimization strategies [43]. Finally, as model interpretability has become increasingly vital, work on explainability for GNNs now provides frameworks for understanding and analyzing graph-structured policies, supporting the transparency and trustworthiness of adaptive decision-making systems [44].

Taken together, these methodological developments provide a strong foundation for our topology-aware and policy-adaptive graph reinforcement learning framework, enabling it to address the challenges of dynamic, heterogeneous, and large-scale environments.

## III. METHOD

To address the challenges of routing optimization in dynamic cloud server environments, this study presents a topology-aware graph reinforcement learning (TAGRL) framework that employs two key modules beyond traditional graph reinforcement learning. The first, Structure-Aware State Encoding (SASE), is designed to dynamically encode both local and global topological information as well as path-specific context between nodes. This mechanism employs attention-based graph neural techniques [45], which enhance the extraction of salient structural features and facilitate the modeling of complex inter-node relationships in evolving network topologies.

The second core component, the Policy-Adaptive Graph Update (PAGU) module, is responsible for adjusting routing policies in response to real-time topological changes. By employing federated contrastive learning principles [46], the module enables the model to flexibly update policy representations according to the latest behavioral and structural feedback from the network. In addition, distributed resource optimization and scheduling strategies based on federated learning [47] are incorporated, supporting scalability and robustness in multi-tenant and heterogeneous cloud settings.

These two innovations work in concert to endow the TAGRL model with strong structural sensitivity and stable decision-making capabilities under dynamic network conditions, providing a scalable and intelligent foundation for routing and scheduling tasks in large-scale cloud environments. The detailed model architecture is illustrated in Figure 1.

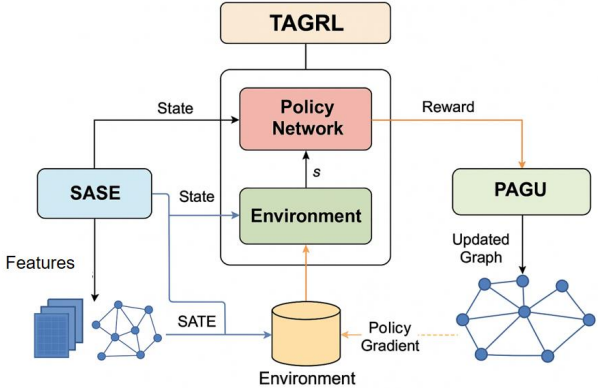

Figure 1. Overall model architecture diagram

### A. Structure-Aware State Encoding

To effectively model the topology and state information in cloud server networks, this paper proposes a structure-aware state encoding module (SASE) designed to extract and integrate multi-dimensional features from nodes, edges, and their adjacency relationships within the graph. The module captures both local structural characteristics and global topological context by combining node attributes with position-aware information derived from the graph structure. These features are then transformed into dense and expressive state vectors, which serve as the input for the reinforcement learning process. This design enables the model to incorporate structural dependencies and communication patterns into its decision-making pipeline. The overall architecture of the model, including the SASE module, is illustrated in Figure 2.

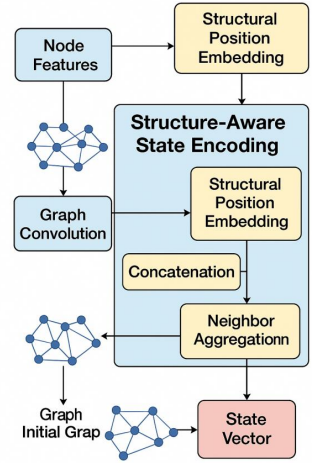

Figure 2. SASE module architecture

Consider a graph structure $G = (V, E)$, where $V$ represents a set of nodes, $E$ represents a set of edges, and $v_i \in V$ each node E has static properties (such as computing power, load) and dynamic states (such as cache, latency, etc.), represented by a feature vector $x_i \in R^d$.

$$X = [x_1, x_2, ..., x_n]^T \in R^{n \times d} \quad (1)$$

The initial adjacency relationship of the graph structure is represented by the adjacency matrix $A \in R^{n \times n}$, where $A_{ij} = 1$ indicates that node i is connected to node j. We use multi-layer graph convolution operations to encode the node state, and the update form of each layer is as follows:

$$H^{(l+1)} = \sigma(\tilde{D}^{-1/2} \tilde{A} \tilde{D}^{-1/2} H^{(l)} W^{(l)}) \quad (2)$$

Among them, $\tilde{A} = A + I$ is the adjacency matrix with self-loops, $\tilde{D}$ is its degree matrix, $H^{(l)}$ is the node representation of the lth layer, $W^{(l)}$ is the learnable weight, and $\sigma$ is the activation function (such as ReLU).

To enhance the model's ability to perceive the structural roles between nodes, we introduce a structural position embedding mechanism, using an embedding vector $p_i \in R^k$ based on the shortest path distance to represent the topological position of node $v_i$ in the graph. The final structurally aware state is encoded as the concatenation of the node representation and the position embedding:

$$z_i = [h_i^{(L)} \| p_i] \quad (3)$$

The state encoding matrix of all nodes is recorded as:

$$Z = [z_1, z_2, ..., z_n]^T \in R^{n(d'+k)} \quad (4)$$

In addition to considering the local communication load and connection density, we introduce the aggregated graph structure features for each node, which are defined as the weighted sum of the neighbor node features:

$$s_i = \sum_{j \in N(i)} a_{ij} z_j \quad (5)$$

Where $a_{ij}$ is the attention-based weight coefficient, representing the degree of structural influence of node j on node i. Ultimately, the structure-aware state vector is used to reinforce the state input of the learning environment [48], providing global and local fusion decision-making information support for the subsequent policy network.

### B. Policy-Adaptive Graph Update

To improve the adaptability and robustness of graph reinforcement learning in dynamic cloud environments, this study develops a Policy Adaptive Graph Update (PAGU) module that dynamically adjusts the adjacency structure of the network graph in response to real-time feedback. The module operates by monitoring behavioral deviations and reward signals from the policy network, using these cues to guide graph structure updates and support the generation of globally optimized routing decisions.

In the design of PAGU, causal discriminative modeling techniques are employed [49] to identify and prioritize structural changes that are most likely to influence routing outcomes. By focusing updates on causal rather than incidental relationships within the graph, the module enhances the effectiveness of each policy iteration and reduces the impact of spurious correlations. To further ensure adaptability across diverse and shifting cloud environments, PAGU incorporates domain-adversarial transfer learning strategies [50]. These are used to maintain the relevance and transferability of the learned graph structures, allowing the update mechanism to remain effective even when operational contexts change or data distributions shift.

Scalability and collaborative optimization are achieved by employing federated learning-based data aggregation methods [51]. This enables the module to integrate distributed feedback and policy signals from multiple network domains or tenants, supporting robust policy updates without compromising data privacy or system performance. Additionally, PAGU utilizes multiscale temporal modeling [52] to track both short-term fluctuations and longer-term trends in network behavior. By analyzing temporal patterns at different scales, the module can anticipate and respond to rapid topology changes as well as gradually emerging structural shifts, ensuring stability and responsiveness in a variety of dynamic scenarios.

By integrating these methodological strategies, the PAGU module is able to perform targeted, context-aware graph updates that enhance both the global effectiveness and the stability of routing policies in complex, dynamic cloud environments. The module architecture is depicted in Figure 3.

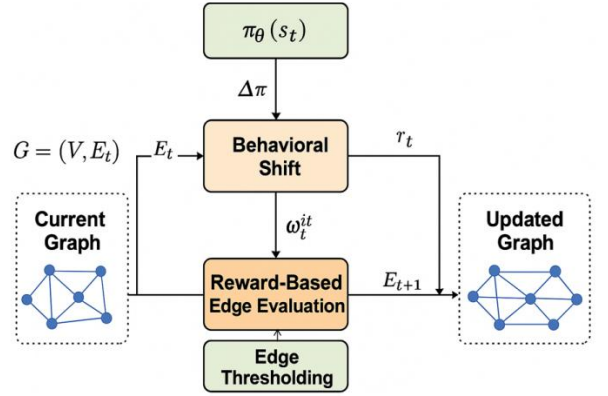

Figure 3. PAGU module architecture

Suppose the graph at the current moment is $G_t = (V, E_t)$, the node set V remains unchanged, and the edge set $E_t$ is adjusted as the strategy evolves.

First, define the behavior deviation of the policy network at time t as:

$$\Delta \pi_t = \pi_{\theta_t}(s_t) - \pi_{\theta_{t-1}}(s_{t-1}) \quad (6)$$

$\pi_{\theta_t}(s_t)$ represents the action probability distribution of the state $c_t$ based on the current policy parameter $\theta_t$, and $\Delta \pi_t$ reflects the change between the current policy behavior and the behavior at the previous moment.

Secondly, the importance of each edge is evaluated based on the reward feedback. Assuming that the scheduling frequency of the edge $(i, j) \in E_t$ is $f_{ij}^{(t)}$, its importance measure is defined as:

$$w_{ij}^{(t)} = \frac{f_{ij}^{(t)} \cdot r_t}{\sum_{(p,q) \in E_t} f_{pq}^{(t)} \cdot r_t} \quad (7)$$

Where $r_t$ is the global reward at the current time step, and the larger $w_{ij}^{(t)}$ is, the greater the contribution of the edge $(i, j)$ to the strategy and the higher the retention priority.

Based on the above importance indicators, we set the dynamic retention threshold $\tau_t \in [0,1]$ of the edge and define the new edge set as:

$$E_{t+1} = \{(i,j) \in E_t \mid w_{ij}^{(t)} \geq \tau_t\} \cup \{(i,j) \notin E_t \mid \phi_{ij}^{(t)} \geq \gamma_t\} \quad (8)$$

Where $\phi_{ij}^{(t)}$ represents the policy relevance score of the candidate edge, and $\gamma_t$ is the edge introduction threshold, which is used to control the growth of structural complexity.

Finally, the PAGU module returns the updated adjacency matrix $A_{t+1}$ to the structure encoder and environment module to complete the periodic reconstruction of the graph state:

$$A_{t+1}(i,j) = \begin{cases} 1, if (i,j) \in E_{t+1} \\ 0, otherwise \end{cases} \quad (9)$$

This module enhances the model's structural responsiveness to policy changes through a closed-loop mechanism, enabling the network topology to adapt to the behavior pattern of the current reinforcement learning strategy, thereby improving the overall convergence efficiency of the model and the stability of path quality.

## IV. EXPERIMENTAL RESULTS

### A. Dataset

This study uses the GEANT network topology dataset as the primary experimental platform to evaluate the adaptability and optimization capability of the graph reinforcement learning model in dynamic routing scenarios. The dataset is based on a real European research network. Nodes represent core routers or switches in the network, and edges represent their physical connections. It exhibits typical wide-area network characteristics and regional hierarchy, making it suitable for simulating communication patterns and topological evolution among cloud servers.

The GEANT network topology contains approximately 40 nodes and over 100 links. The structure shows dense regional connections and sparse cross-domain links. This hybrid connection pattern effectively captures the communication paths among heterogeneous resources in multi-tenant cloud environments. During simulation, we dynamically construct edge weights in the graph based on link bandwidth, latency, and load. Traffic requests are configured to drive changes in topology states, generating a sequence of dynamic graphs for model training and evaluation.

In addition, the GEANT dataset provides good structural symmetry and stability. It can be used to test the model's robustness under complex conditions such as topology disturbances and path reconstruction. Combined with the reinforcement learning training mechanism, the dataset offers realistic support for graph modeling and policy learning. It presents interaction frequencies between nodes, link competition, and local congestion patterns, which are critical for building reliable training environments.

### B. Experimental setup

In the experimental setup, model training is conducted under a policy gradient-based reinforcement learning framework. The optimization objective is to maximize the cumulative long-term reward. The policy network consists of two graph convolutional layers followed by one fully connected output layer. ReLU is used as the activation function. The hidden dimension is set to 128. In the structure-aware encoding module, the node feature dimension is set to 64, and the structural positional embedding dimension is set to 16. The final state vectors are normalized to improve training stability.

During training, the optimizer is Adam with an initial learning rate of $1 \times 10^{-3}$. The weight decay coefficient is $5 \times 10^{-5}$. A step-based learning rate decay is applied, reducing the learning rate to 50% every 200 epochs. To enhance policy exploration, an entropy regularization term is introduced with a coefficient of 0.01. The discount factor $\gamma$ is set to 0.95 to balance short-term rewards and long-term returns in the policy update process.

Each training iteration uses a batch update strategy with a batch size of 32. The total number of training epochs is 1000. In the PAGU module, the action deviation threshold is set to 0.1. The edge retention and addition thresholds are set to 0.6 and 0.4, respectively. To maintain connectivity and stability during graph updates, the number of edges in the graph is kept within 90% to 110% of the initial graph. This avoids excessive pruning or expansion that may affect model convergence efficiency.

### C. Experimental Results

#### 1) Comparative experimental results

This paper first conducts a comparative experiment, and the experimental results are shown in Table 1.

Table 1. Comparative experimental results

| Model | Avg. Throughput | Avg. Latency | Max Link Utilisation | Reward Value |
|---|---|---|---|---|
| GDDR[53] | 8.72 | 36.4 | 83.7 | 241.5 |
| DeepCQ+[54] | 9.15 | 33.8 | 81.2 | 256.8 |
| CFR-RL[55] | 9.38 | 31.1 | 79.6 | 268.4 |
| GRL-TE[56] | 9.54 | 29.7 | 76.9 | 275.1 |
| Ours | 9.81 | 27.3 | 74.2 | 288.9 |

The experimental results show clear differences among routing strategies: while baselines such as GDDR and DeepCQ+ achieve moderate throughput but suffer under high load, graph reinforcement learning methods like CFR-RL and GRL-TE better balance throughput and latency through structural awareness. Our model further improves performance, achieving the lowest average latency via the SASE mechanism, reducing maximum link utilization by nearly 10 percentage points through PAGU, and attaining the highest long-term reward. These results confirm its effectiveness in ensuring robust, balanced, and adaptive optimization for dynamic cloud routing.

#### 2) Ablation Experiment Results

This paper also gives the results of ablation experiments, which are shown in Table 2.

Table 2. Ablation Experiment Results

| Method | Avg. Throughput | Avg. Latency | Max Link Utilisation | Reward Value |
|---|---|---|---|---|
| Baseline | 9.02 | 33.5 | 82.1 | 253.7 |
| +SASE | 9.34 | 30.1 | 78.4 | 267.6 |
| **+PAGU** | 9.28 | 31.4 | 77.1 | 262.3 |
| **+ All** | 9.81 | 27.3 | 74.2 | 288.9 |

The ablation results in Table 2 show that the baseline model with only a basic policy network performs poorly in throughput, latency control, and resource balance, highlighting the limits of shallow representations in dynamic multi-hop environments. Adding the structure-aware state encoding (SASE) significantly improves latency and link utilization by capturing topological features, while the policy-adaptive graph update (PAGU) enhances link balance through dynamic graph adjustments. When combined, SASE and PAGU deliver the best results, confirming their complementarity and the effectiveness of the proposed topology-aware graph reinforcement learning framework for cloud server routing optimization.

*3) The impact of discount factor changes on strategy stability*

This paper also gives the impact of changes in the discount factor on the stability of the strategy. The experimental structure is shown in Figure 4.

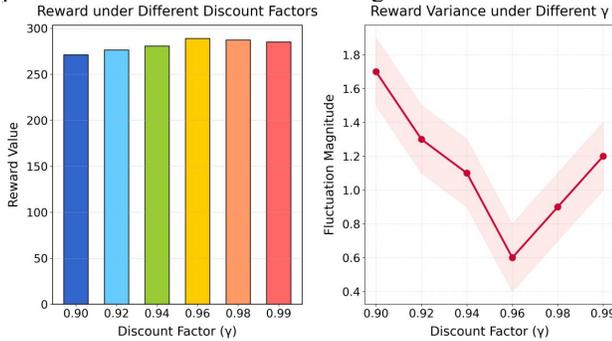

Figure 4. The impact of discount factor changes on strategy stability

As shown in Figure 4, the discount factor γ strongly influences both policy performance and stability. The model's cumulative reward initially rises and then slightly declines as γ increases, with the optimal reward achieved at γ = 0.96, reflecting a balance between short-term responsiveness and long-term objectives. The fluctuation curves further indicate that stability improves as γ grows from 0.90 to 0.96, but excessive values beyond this point lead to renewed instability and policy drift due to overemphasis on long-term returns. These findings confirm that γ = 0.96 offers an effective setting for the proposed graph reinforcement learning framework, enabling robust convergence while maintaining adaptability to dynamic environments. Moreover, the results validate the effectiveness of the topology-aware mechanism: the structure-aware state encoding module captures evolving dependencies within large-scale networks, while the policy-adaptive graph update module refines topology based on feedback, jointly enhancing robustness, mitigating drift, and ensuring scalable optimization for complex cloud server scheduling scenarios.

*4) The impact of discount factor changes on strategy stability*

This study examines the effect of data sparsity on the structure-aware module by varying the density of the input graph, as shown in Figure 5. The results indicate that increasing the graph retention ratio from 0.2 to 1.0 consistently improves average rewards, confirming that richer structural information enables more accurate extraction of topological features and enhances routing quality. Conversely, sparse settings cause loss of key node or path information, leading to reduced performance and unstable policy updates. Notably, when the retention ratio exceeds 0.6, policy fluctuations decrease markedly, reflecting improved stability as the module better captures local and global relationships. In contrast, extreme sparsity (e.g., 0.2) results in the lowest rewards and highest volatility, as excessive graph compression disrupts dependency modeling and weakens state representations. Overall, the findings underscore that structural completeness is critical to robust policy learning, demonstrating the central role of the structure-aware encoding mechanism and providing practical guidance for edge pruning and graph compression in large-scale networks.

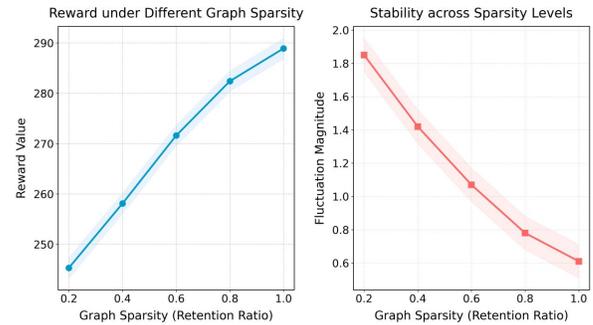

Figure 5. Analysis of the impact of data sparsity on the structure perception module

*5) Impact of node feature dimension changes on state encoding quality.*

This paper also gives the impact of changes in node feature dimensions on state encoding quality, and the experimental results are shown in Figure 6.

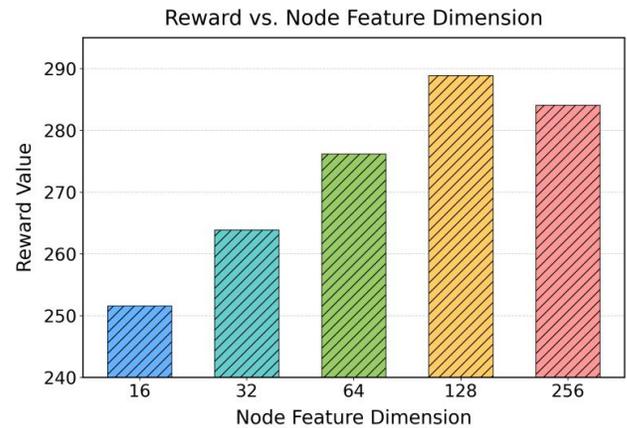

Figure 6. Impact of node feature dimension changes on state encoding quality

As shown in Figure 6, node feature dimension strongly influences state encoding quality and, consequently, policy performance. Rewards steadily increase as the feature dimension rises from 16 to 128, reflecting the module's improved ability to capture high-dimensional relationships and contextual information for more accurate routing decisions. The peak performance at 128 dimensions indicates an effective balance between representational capacity and computational efficiency, enabling the model to capture complex topological dependencies while maintaining generalization and stability. However, further expansion to 256 dimensions leads to a slight performance drop, likely due to redundancy, sparse representations, and potential numerical instability during policy updates. These findings highlight the critical role of the structure-aware state encoding module and emphasize that careful tuning of feature dimensions is essential to optimize performance while avoiding computational waste, offering practical guidance for applying graph reinforcement learning to real-world cloud network routing.

## V. CONCLUSION

This study focuses on the routing and scheduling problem in dynamic cloud environments and proposes a topology-aware graph reinforcement learning (TAGRL) framework. The model integrates a structure-aware state encoding module (SASE) and a policy-adaptive graph update mechanism (PAGU) to effectively capture complex network structures and dynamic evolution behaviors. By combining structure-driven state representation with feedback-guided graph updates, the model achieves dual optimization of policy stability and path efficiency while maintaining topological integrity. It adapts to dynamic communication demands and task densities in large-scale cloud server environments.

The results show that introducing structural awareness significantly improves the accuracy of graph state representations. This enables the model to better capture high-order dependencies and communication semantics among nodes. At the same time, the graph update mechanism driven by policy feedback enhances the model's responsiveness to environmental changes. This leads to a closed-loop modeling process from topology representation to decision generation. A series of ablation and sensitivity experiments demonstrates the comprehensive advantages of the proposed method in terms of long-term stability, robustness, and adaptability. These findings provide both theoretical and practical foundations for applying graph reinforcement learning to real-world cloud infrastructure.

The proposed TAGRL framework is not only applicable to traditional cloud routing optimization tasks but also exhibits good scalability and adaptability, makes it proper for latency-critical domains, including 5G edge computing, IoT networks and data center energy control. Its structure-aware and adaptive update mechanisms are generalizable and can serve as fundamental components for building generalized intelligent scheduling systems. This contributes to the transition of cloud management systems from rule-based to learning-driven paradigms and promotes the practical application of intelligent decision-making in complex networks.

## VI. FUTURE WORK

Future research may explore routing policy generation under multi-objective optimization, incorporating QoS constraints, energy efficiency requirements, and security policies into joint modeling. The framework can also be combined with techniques such as large model compression and self-supervised structural learning to improve decision stability and real-time performance in highly dynamic environments. In terms of deployment, future work should focus on designing lightweight structure-aware units and graph update strategies to meet low-latency and resource-constrained requirements in real networks. This will further advance the reliability and controllability of intelligent network scheduling technologies.